\pdfoutput=1
\documentclass[review]{elsarticle}
\usepackage{hyperref}
\usepackage{multirow} 
\usepackage{booktabs}
\usepackage{makecell}
\usepackage{graphicx}
\usepackage{amsfonts} 
\usepackage{amsmath}
\usepackage{array}
\usepackage{url}
\usepackage{longtable}
\usepackage{rotating}
\usepackage[normalem]{ulem}
\usepackage[linewidth=1pt]{mdframed}

\biboptions{sort&compress}
\allowdisplaybreaks[4]

\usepackage{lineno}

\journal{Information Fusion}

\begin{document}

\begin{frontmatter}

\title{MESSFN : a Multi-level and Enhanced Spectral-Spatial Fusion Network for Pan-sharpening}


\author[mymainaddress]{Yuan Yuan}
\ead{y.yuan1.ieee@gmail.com} 

\author[mymainaddress,mysecondaryaddress]{Yi Sun}
\ead{yisun98nwpu@gmail.com} 

\author[mymainaddress]{Yuanlin Zhang \corref{mycorrespondingauthor}}
\cortext[mycorrespondingauthor]{Corresponding author}
\ead{zhangyuanlin.ucas@gmail.com}

\address[mymainaddress]{School of Artificial Intelligence, Optics and Electronics (iOPEN), Northwestern Polytechnical University, Xi’an 710072, China}
\address[mysecondaryaddress]{School of Computer Science, Northwestern Polytechnical University, Xi’an 710072, China}

\begin{abstract}
Through fusing low resolution \textit{Multi-Spectral} (MS) images and high resolution \textit{PANchromatic} (PAN) images, pan-sharpening provides a satisfactory solution for the demands of high resolution multi-spectral images. 
However, dominant pan-sharpening frameworks simply concatenate the MS stream and the PAN stream once at a specific level. This way of fusion neglects the multi-level spectral-spatial correlation between the two streams, which is vital to improving the fusion performance.
In consideration of this, we propose a \textit{Multi-level} \textit{and} \textit{Enhanced} \textit{Spectral-Spatial} \textit{Fusion Network} (MESSFN) with the following innovations: 
First, to fully exploit and strengthen the above correlation, a \textit{Hierarchical Multi-level Fusion Architecture} (HMFA) is carefully designed.
 A novel \textit{Spectral-Spatial} (SS) stream is established to hierarchically derive and fuse the multi-level prior spectral and spatial expertise from the MS stream and the PAN stream. 
This helps the SS stream master a joint spectral-spatial representation in the hierarchical network for better modeling the fusion relationship. 
Second, to provide superior expertise, consequently, based on the intrinsic characteristics of the MS image and the PAN image, two feature extraction blocks are specially developed.
 In the MS stream, a \textit{Residual Spectral Attention Block} (RSAB) is proposed to mine the potential spectral correlations between different spectra of the MS image through adjacent cross-spectrum interaction.  While in the PAN stream, a \textit{Residual Multi-scale Spatial Attention Block} (RMSAB) is proposed to capture multi-scale information and reconstruct precise high-frequency details from the PAN image through an improved spatial attention-based inception structure. The spectral and spatial feature representations are enhanced.
Extensive experiments on two datasets demonstrate that the proposed network is competitive with or better than state-of-the-art methods. Our code can be found in \href{https://github.com/yisun98/MESSFN}{github}\footnote{\href{https://github.com/yisun98/MESSFN}{https://github.com/yisun98/MESSFN}}.

\end{abstract}

\end{frontmatter}

\section{Introduction}
Due to the limitations of the physical equipment, it’s difficult to obtain both high spectral resolution and high spatial resolution of remote sensing images. The satellite sensors usually capture two types of images: \textit{PANchromatic} (PAN) images with high spatial resolution but low spectral resolution, and \textit{Multi-Spectral} (MS) images with high spectral resolution but low spatial resolution. 
\begin{figure}[htbp]
	\centering
	\includegraphics[width=\linewidth]{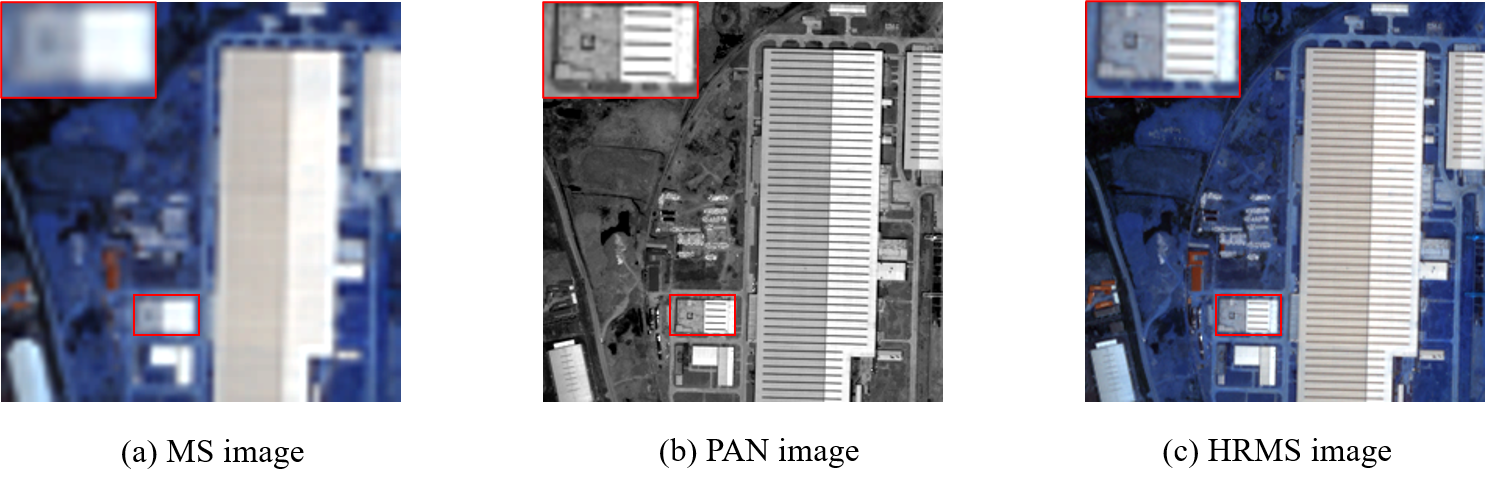}
	\caption{The schematic diagram of pan-sharpening. (a) MS image (re-sampled) (b) PAN image (c) HRMS image.}
	\label{fig-pansharpening}
\end{figure}
In recent years, significant breakthroughs have been made in remote sensing technology, which has improved the speed and quality of imaging in terms of spectral or spatial. However, it’s still insufficient for MS images and PAN images to meet both high spectral resolution and high spatial resolution demands of practical applications, such as urban planning \cite{Lazaridou2016}, land-cover classification \cite{Gaetano2018}, and environmental monitoring \cite{Li2018a}. As shown in Fig.\ref{fig-pansharpening}, pan-sharpening provides a satisfactory solution through fusing the spectral information of the MS image and the spatial information of the PAN image to generate a \textit{High} spatial \textit{Resolution Multi-Spectral} (HRMS) image, which is conducive to high-level analysis and decision-making.
\begin{figure}[htbp]
	\centering
	\includegraphics[width=\linewidth]{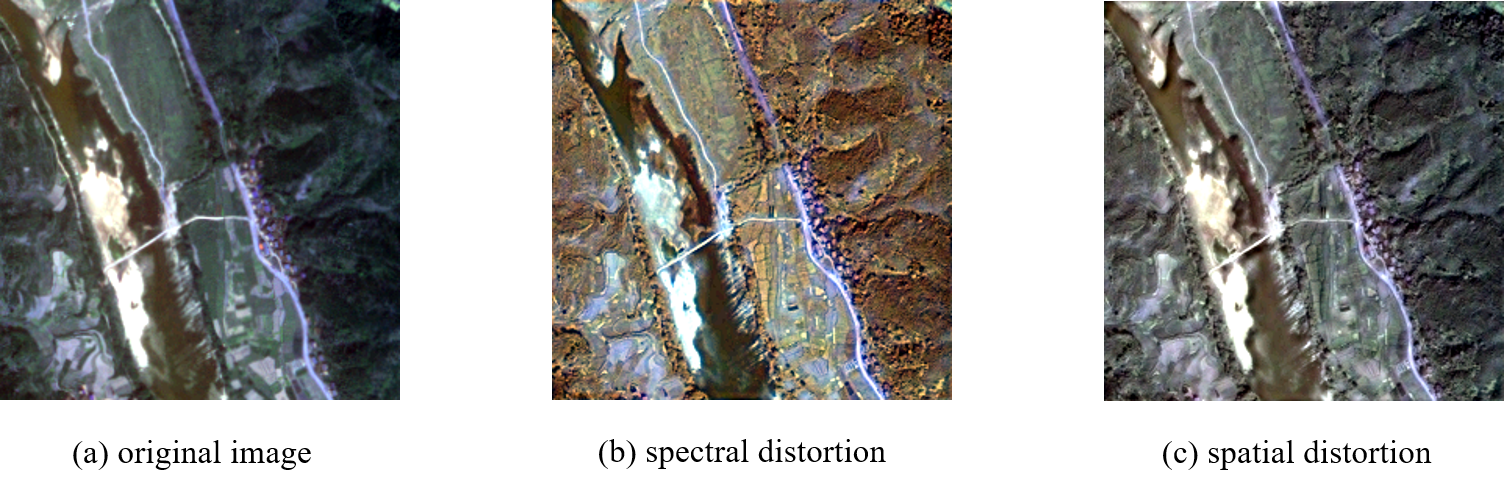}
	\caption{Examples of the distortions. (a) original image (b) spectral distortion (c) spatial distortion.}
	\label{fig-distortion}
\end{figure}

Over the past few decades, great efforts have been devoted to the pan-sharpening task and many classical methods \cite{Li2016, Sowmya2017,Ayas2017} have been proposed. However, the linear models usually cause serious spectral or spatial distortion in the fusion results, as shown in Fig.\ref{fig-distortion}.
In recent years, \textit{Convolutional Neural Networks} (CNNs) have shined with their astonishing nonlinear learning ability
in a series of visual tasks, such as image classification \cite{Zhang2020}, image segmentation \cite{Shelhamer2016}, and object detection \cite{Zhang2019}. This inspired researchers to introduce CNNs to the pan-sharpening task.
Masi \textit{et al.} \cite{Masi2016} successfully used SRCNN \cite{Dong2014} model and achieved better results than traditional methods. Yang \textit{et al.} \cite{Yang2017} followed the single-input image super-resolution structure and trained a deep residual network in the high-pass domain. Shao and Cai \cite{Shao2018} suggested using two branches to take full advantage of the characteristics of the two input images. Liu \textit{et al.} \cite{Liu2019} proposed a two-stream architecture and two sub-networks were constructed to encode the source images separately.
However, careful observation of the fusion results reveals that they are still suffering from spectral deviations and spatial blurring. Therefore, the HRMS images don’t achieve the ideal quality.

\begin{figure}[htbp]
	\centering
	\includegraphics[width=\linewidth]{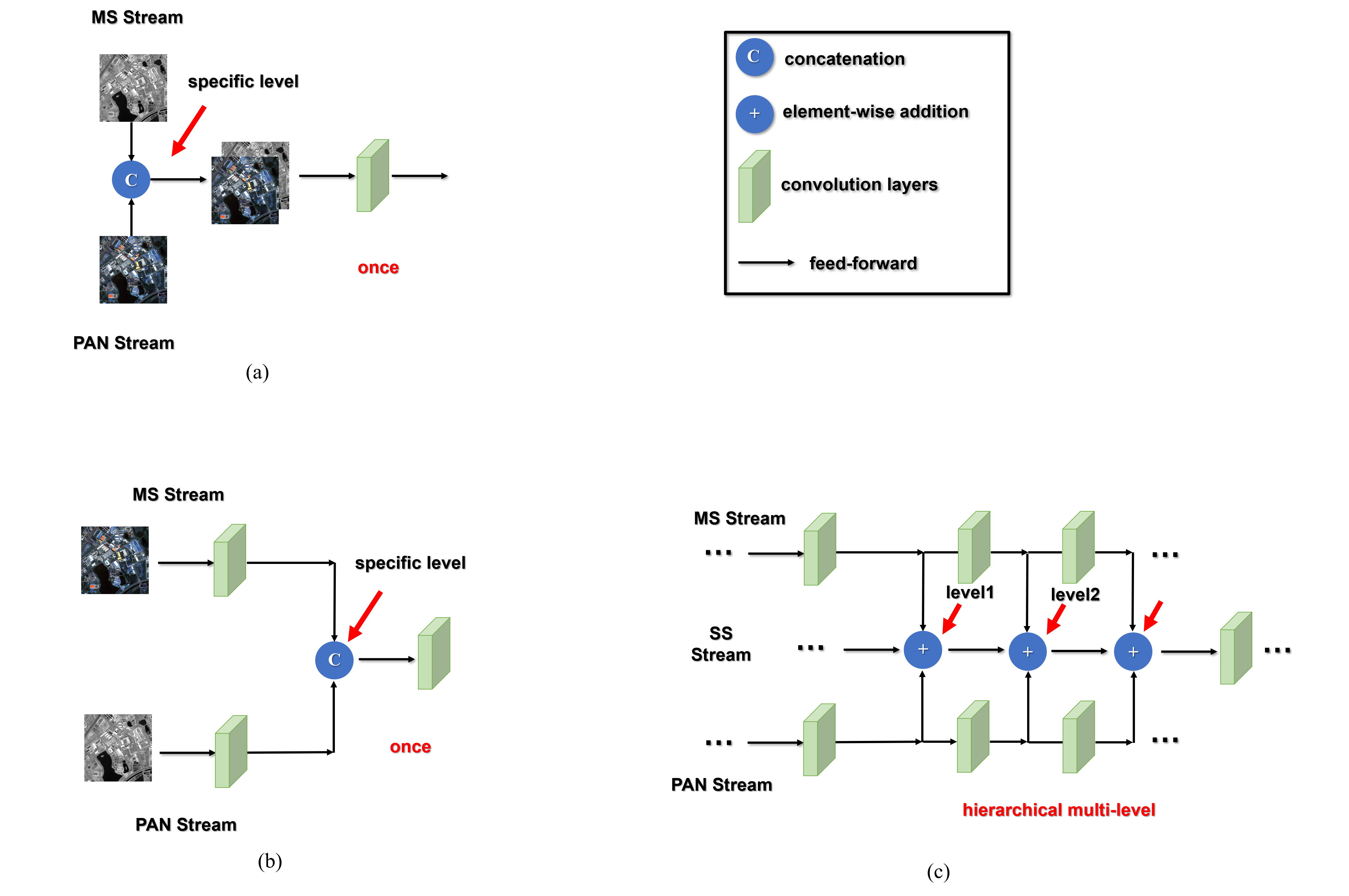}
	\caption{Comparison of pan-sharpening prototypes. The red arrow indicates the fusion level of the spectral and spatial. (a) one-stream prototype (b) two-stream prototype (c) multi-level fusion prototype (\textbf{ours}), and the third SS stream is the joint Spectral-Spatial (SS) representation.}   
	\label{fig-prototypes}
\end{figure}

Two design prototypes (one-stream and two-stream) shown in Fig. \ref{fig-prototypes} (a) and (b) are available through analyzing the existing dominant fusion frameworks. 
However, the networks based on these prototypes (e.g., PNN \cite{Masi2016}, RSIFNN \cite{Shao2018} ) are limited in fusion performance. The reason for this is that the MS stream and the PAN stream are simply concatenated and fused in these networks 
once in a specific level [as shown by the red arrow in Fig. \ref{fig-prototypes} (a) and (b)]. 
This way of fusion neglects the multi-level spectral-spatial correlation between the spectral information in the MS stream and the spatial information in the PAN stream at other levels
in the hierarchical network. 
Importantly, this correlation is crucial to tune the representation of spectral and spatial features, reduce the redundancy of spectral and spatial information, thereby improving the fusion performance of the network and obtaining high-quality fusion results.

In consideration of this, in this paper, we propose a \textit{Multi-level} and \textit{Enhanced} \textit{Spectral-Spatial} \textit{ Fusion Network} (MESSFN), as illustrated in Fig. \ref{fig-framework}. First, 
to strengthen the above correlation, a \textit{Hierarchical Multi-level Fusion Architecture} (HMFA) is carefully designed in the network. Specifically, as shown in Fig. \ref{fig-prototypes} (c), a novel \textit{Spectral-Spatial} (SS) stream is established parallel to the MS stream and the PAN stream. Note that with the optimization of the network, the MS stream and the PAN stream can learn the expert spectral and spatial feature representations from the MS image and the PAN image, respectively. Therefore, the SS stream can hierarchically derive and fuse the multi-level prior spectral and spatial expertise from the MS stream and the PAN stream. This not only helps the SS stream effectively master a joint spectral-spatial representation in the hierarchical network for better modeling the fusion relationship but also greatly suppresses the spectral and spatial distortions. 

Second, to provide superior expertise, two feature extraction blocks are specially developed based on the intrinsic characteristics of the MS image and the PAN image. 
Specifically, in the MS stream, a \textit{Residual Spectral Attention Block} (RSAB) is proposed. Through adjacent cross-spectrum interaction along the spectral dimension, the potential correlations between different spectra of the MS image can be fully mined. Spectral process and prediction are enhanced to prevent spectral deviations. 
In the PAN stream, a \textit{Residual Multi-scale Spatial Attention Block} (RMSAB) is proposed to capture multi-scale information using inception structure, and an improved spatial attention is designed to reconstruct precise high-frequency details from the PAN image. The spatial quality can be effectively enhanced.

In summary, the contributions of our work can be summarized as follows:
\begin{enumerate}[(1)]
\item We propose an effective pan-sharpening framework MESSFN to fully strengthen the multi-level spectral-spatial correlation. Extensive experiments on the WorldView-II and GaoFen-2 datasets demonstrate that MESSFN achieves a pleasing dual-fidelity in the spectral and spatial domains.

\item We carefully design a HMFA. The SS stream hierarchically derives and fuses multi-level prior spectral and spatial expertise from the MS stream and the PAN stream. The SS stream can master a joint spectral-spatial representation in the hierarchical network for better modeling the fusion relationship.

\item To provide superior expertise, RSAB and RMSAB are proposed based on the intrinsic characteristics of the source images. RSAB is exploited to mine the potential correlations between different spectra of the MS image through adjacent cross-spectrum interaction. RMSAB is developed to capture multi-scale information and reconstruct precise high-frequency details from the PAN image. The spectral and spatial feature representations are enhanced.

\end{enumerate}

The remaining structure of the article is illustrated as follows: Section \ref{sec-relatedwork} is Related Works. We briefly review the previous pan-sharpening work. Section \ref{sec-proposedmethod} is Proposed Method. We introduce the proposed MESSFN in detail, including the framework, the hierarchical multi-level fusion architecture, and the proposed blocks. The experimental demonstration and analysis are provided in Section \ref{sec-experiments}, we also discuss some impact factors of the fusion performance. The conclusions are presented in Section \ref{sec-conclusion}. 

\section{Related Works}
\label{sec-relatedwork}
\subsection{Pan-sharpening Based on Traditional Methods}
Traditional methods consist of three types:\textit{ Component Substitution} (CS), \textit{Multi-Resolution Analysis} (MRA), and\textit{ Model Optimization} (MO).
The primary idea of the CS methods is to separate the spectral component and the spatial component of the MS image through linear space transformation, and use the spatial component extracted from the PAN image to perform the component substitution, and finally inverse the transformation to the original space. Such methods suffer from spectral distortion because of the inconsistent bandwidth of the above spatial components. Typical algorithms include \textit{Principal Component Analysis} (PCA) \cite{Shah2007a}, \textit{Gram-Schmidt} (GS)  \cite{Li2016}, \textit{Intensity-Hue-Saturation} (IHS) \cite{Rahmani2010}, etc.
The MRA methods first use multi-scale tools to decompose the source images into sub-signals with different frequencies and resolutions, and then execute specific fusion rules on the corresponding levels of sub-signals, and finally reconstruct from the handled sub-signals. However, such methods are influenced heavily by the decomposition level and fusion rules. More studies are available on the MRA methods, most of which are based on various wavelet transforms \cite{2002Multi, Kallel2015, Sowmya2017, Shah2007, Upla2014}, and pyramid transforms\cite{Miao2007,BARONTI2003}.
In the MO methods, sparse representation theory \cite{Ayas2017} {\tiny }emerges rapidly and more improvements are attached to this theory \cite{Chen2015, Chen2018, Dalla2015}. It constructs a joint dictionary from the low resolution and the high resolution image pairs, and solves the sparse coefficients to obtain the sparse representation of the source images. However, this theory represents the image signal with a simple linear combination of a few dictionary atoms, which limits the expression ability of the model.

\subsection{Pan-sharpening Based on CNNs}
Inspired by the super resolution model SRCNN \cite{Dong2014}, Masi \textit{et al.} \cite{Masi2016} followed the same architecture and used a three-layer network for pan-sharpening. After achieving better results than traditional methods, a new generation of fusion boom was sparked. Yang \textit{et al.} \cite{Yang2017} incorporated domain expertise into the deep learning framework and trained the network in the high-pass domain instead of the image domain, and the performance reached the state-of-the-art at that time. Rao \textit{et al.} \cite{Rao2017} considered that the difference between the MS image and HRMS image is spatial information, which can be obtained from the PAN image. Afterward, Wei \textit{et al.} \cite{Wei2017} built a deep residual network and introduced a long-distance skip connection to stand for residual learning in the first fusion stage. Liu \textit{et al}. \cite{Liu2019}  proposed a two-stream architecture, which altered the previous way of stacking bands as input and obtained the HRMS image in the feature level. Shao and Cai \cite{Shao2018}  suggested using two branches to extract features from the two input images respectively. They emphasized more on the spatial information, therefore selecting a deeper branch on the PAN image than the MS image. Liu \textit{et al.} \cite{Liu2020} explored the Generative Adversarial Network (GAN) architecture to this task and drove the advances in the fusion of GAN \cite{Liu2020, Shao2020, Ozcelik2020, Ma2020}. Prominently, Ma \textit{et al.} \cite{Ma2020} proposed an unsupervised framework to tackle the supervised training strategy. The spectral discriminators and the spatial discriminators are established to approximate the generated HRMS image consistent with the source images distribution. Besides, there are some other fusion models, such as the dense connection in DenseNet \cite{ Wang2019, Peng2020} the progressive up-sampling method \cite{Liu2020}, the feedback
mechanism \cite{Fu2020}, \textit{etc}. In conclusion, CNNs bring a new exploring direction and a superior fusion performance for pan-sharpening.

\section{Proposed Method}
\label{sec-proposedmethod}

\subsection{Framework}
\begin{figure*}[htbp]
	
		\centering
		\includegraphics[width=0.9\linewidth]{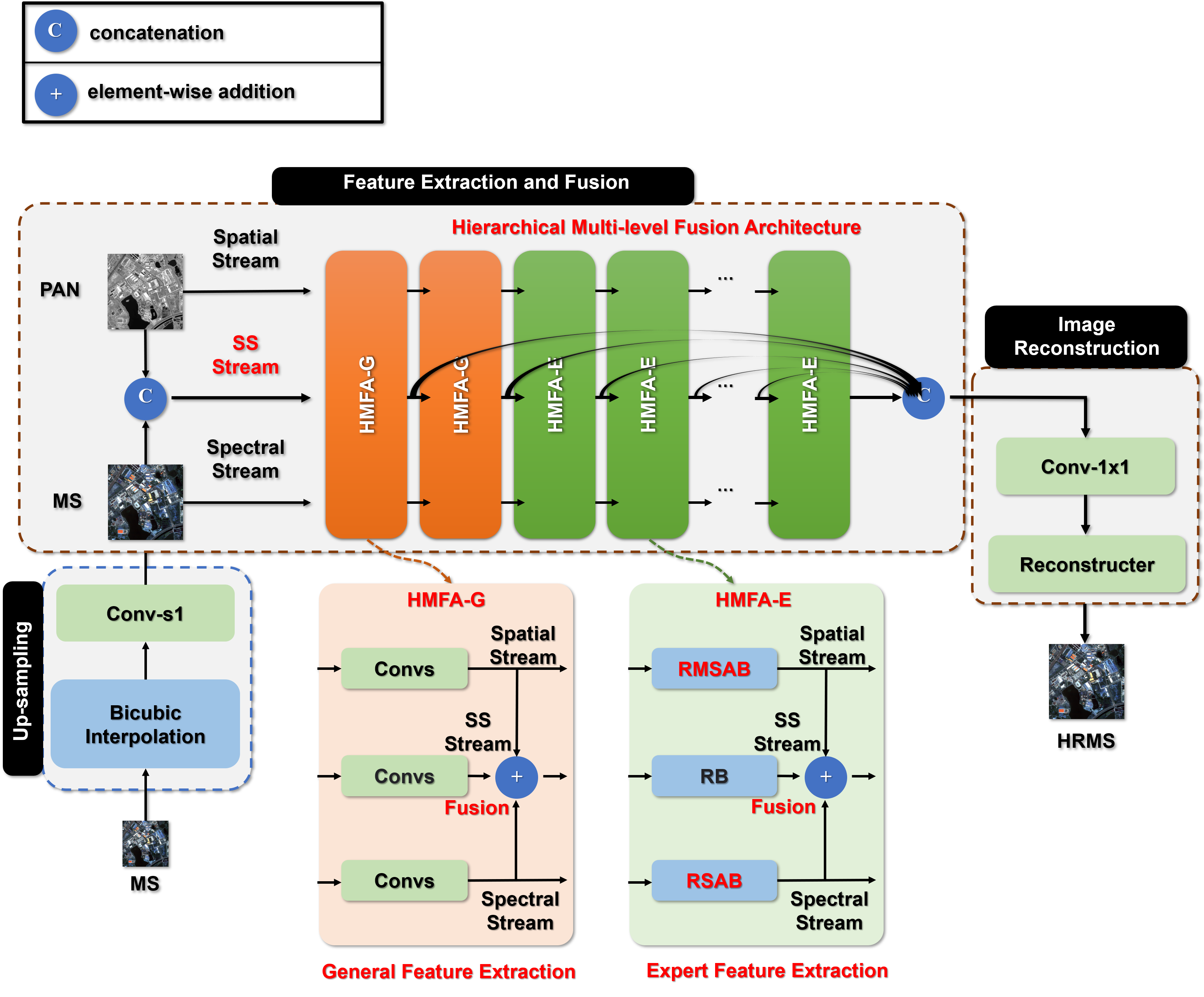}
	\caption{The overall architecture of the proposed MESSFN, which consists of Up-sampling, Feature Extraction and Fusion, and Image Reconstruction.
	  }
	\label{fig-framework}
\end{figure*}
As demonstrated in Fig.\ref{fig-prototypes} (c), in the hierarchical network, the spectral-spatial correlation can be fully exploited and strengthened at different levels. 
In consideration of this, we propose a \textit{Multi-level} \textit{and} \textit{Enhanced} \textit{Spectral-Spatial} \textit{Fusion Network} (MESSFN). In this section, we first give the overall architecture of the proposed MESSFN, which is composed of the following parts: \textit{UP-sampling} (UP), \textit{Feature} \textit{Extraction} \textit{and} \textit{Fusion} (FEF), and \textit{Image} \textit{Reconstruction} (IR), as illustrated in Fig.\ref{fig-framework}.

First, in UP, the MS image is up-sampled to the same size as the PAN image with an optimized strategy. And then based on the hierarchical multi-level fusion architecture, FEF extracts the spectral and spatial features from the source images for general and expert fusion. Finally, IR reconstructs the HRMS image from the feature domain with different levels of features.

\subsubsection{Up-sampling}
The spectral noise introduced in the traditional interpolation methods requires the network to have strong robustness \cite{Lei2021}. And the learnable transposed convolution \cite{Zeiler2011} brings checkerboard artifacts in the image generation problem. Therefore, we introduce   \textit{Bicubic}-\textit{Resize} \textit{Convolution} (BRC) for the first time to optimize the up-sampling process in pan-sharpening. The principle behind it has been described in detail in \cite{Odena2016}. The spatial resolution of the up-sampled MS image is effectively improved at the beginning, which implicitly enhances the subsequent feature extraction of the MS stream and the SS stream.
Denote $ I_{MS} \in \mathbb{R}^{h \times w \times c}$  as the input MS image ($h, w$, and $c$ are the height, width and number of spectral bands of the MS image, respectively), the version of up-sampled MS image $ I_{MS}\uparrow \in \mathbb{R}^{H \times W \times C}$ can be formulated as
\begin{equation}
I_{MS\uparrow} = Conv(Bic ( I_{MS} ))
\end{equation}
where  $Bic(\cdot)$ and $Conv(\cdot)$ are the bicubic interpolation and the convolution with stride is 1, respectively.

\subsubsection{Feature Extraction and Fusion} 
\label{fef}
Convolution is popularly used for general feature extraction in pan-sharpening 
\cite{Tong2017, Lei2021}. 
The main advantages are twofold. 
	On the one hand, convolution is good at early visual feature extraction. This is important to make the training process more stable and achieve the optimal performance \cite{Xiao2021}. 
	On the other hand, convolution gives a simple strategy to map the source image to the high-dimensional feature space. Therefore, we first apply two layers with normal 3 x 3 convolution to extract the general features from the MS image and the PAN image. 
	
The transition from low-level visual features to high-level semantic features, which indicates that the network gradually learns the key feature representation \cite{Johnson2016}. We suggest that the network masters the expertise of the source image. 
We should note that it is important to provide the superior expertise. Therefore, we specially designed the feature extraction blocks of the MS stream and the PAN stream based on the intrinsic characteristics of the MS image and the PAN image. And in the SS stream, we use the standard Residual Block (RB) to extract spectral-spatial features. 
We will describe them in detail later. The general features are sent to the $ B $ cascaded feature extraction blocks for extracting the expert features. In this way, the network can learn rich hierarchical information. 
Since the operation of general and expert fusion is consistent, the output of the ($ k+1 $)-th feature can be expressed as
\begin{equation}
\begin{split}
f_{}^{k+1} &=  H(f_{}^{k})  \\
\end{split}
\end{equation}
where $H(\cdot)$ is the function of feature extraction in FEF.

In addition, it is worth noting that in the general and expert feature extraction and fusion, we introduce different feature extraction functions $H(\cdot)$ (normal convolution and specially designed blocks). The specific reason is consistent with the above description, which is also carefully designed by us.

\subsubsection{Image Reconstruction}
For the final image reconstruction work, there are two
points to be noted: first, different levels of features have different contributions and how to make full use of them is important. Second, it is not sufficient to use only the last layer of abstract features. Extensive works on the super-resolution and the pan-sharpening \cite{Li2018, Hu2019, Peng2020} have verified these considerations. Therefore, we adopt a simple strategy that adaptively aggregates and fuses the features from  different levels using 1 x 1 convolution.
\begin{align}
\begin{split}
& f_g = Conv( (f_{}^{0} \ \copyright \ f_{}^{1} \  ... f_{}^{B+1 } \ \copyright \ f_{}^{B+2}) ) \\
\end{split}
\end{align}
where $f_g$ denotes the feature of adaptive aggregation and fusion, and $\copyright$ is concatenation operation. And $f_{}^{i}$ represents the output feature maps of $i$-th level in the network. These features are employed to reinforce the high-level semantic features and exploit helpful information for obtaining the high-quality fusion results. 

Finally, the prediction image \textit{$O_{HRMS}$} is obtained by 3 x 3 convolution in the feature domain, which can be expressed as
\begin{align}
\begin{split}
& O_{HRMS} = \tau (Conv( f_{g})) 
\end{split}
\end{align}
where $\tau$ denotes the Tanh activation function.

\subsection{Hierarchical Multi-level Fusion Architecture}
The feature extraction and fusion are based on the \textit{Hierarchical Multi-level Fusion Architecture} (HMFA). Specifically, in the HMFA, a novel \textit{Spectral-Spatial} (SS) stream is established parallel to the MS stream and the PAN stream. The fundamental prototype is formed. Different from the previous pan-sharpening frameworks, HMFA focuses on how to fully exploit and strengthen the multi-level spectral-spatial correlation. Therefore we carefully design this architecture. 

First, as described in Section \ref{fef}, with the optimization of the proposed MESSFN, the MS stream and the PAN stream can master the spectral expertise and the spatial expertise, respectively. 

\begin{figure}[htbp]
	\centering
	\includegraphics[width=\linewidth]{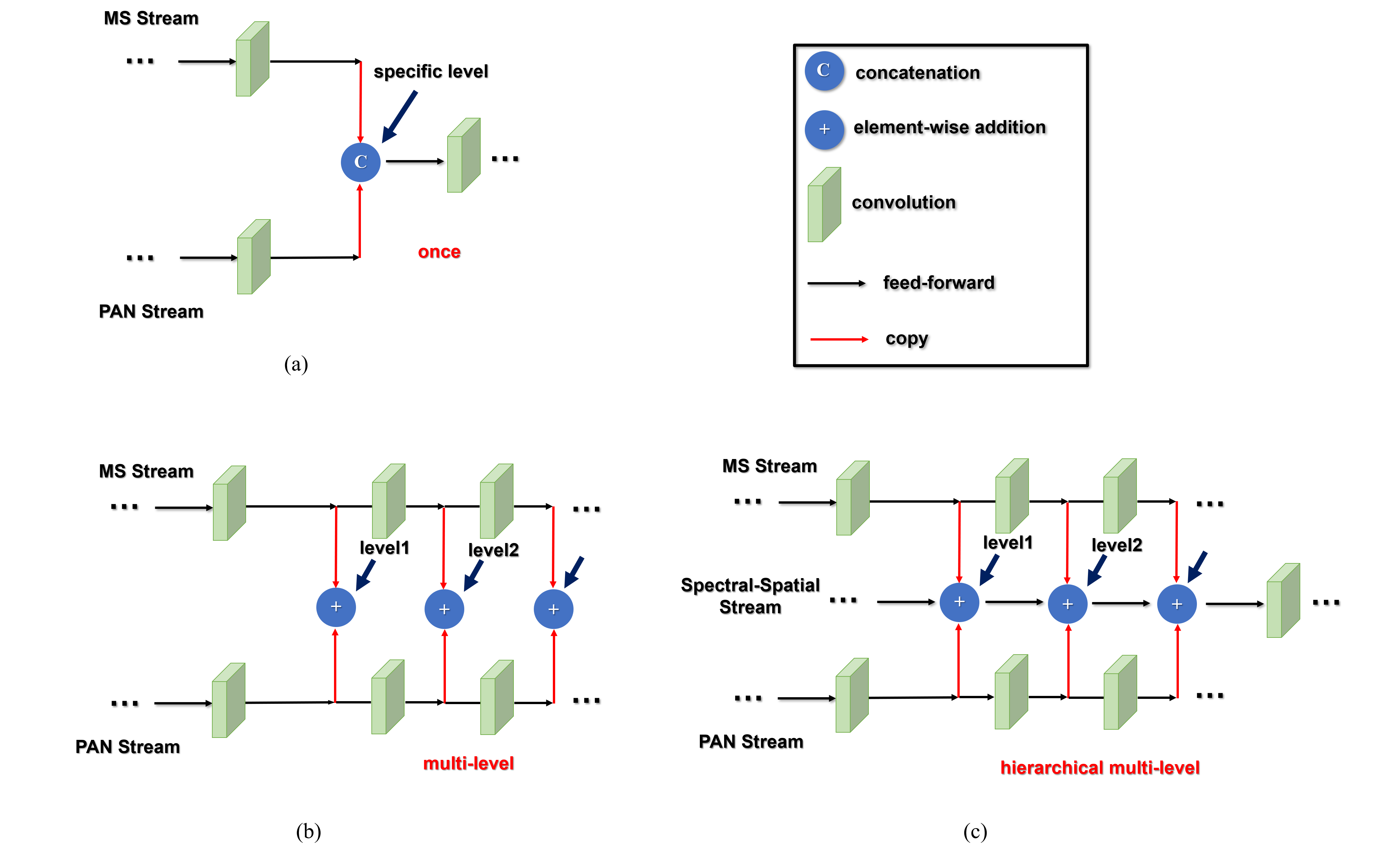}
	\caption{Comparison of the fusion architecture. (a) general fusion (b) multi-level fusion (c) hierarchical multi-level fusion. (a) is the previous general architecture, (b) and (c) are the proposed architecture in this paper, and (c) is the advanced version of (b).}
	\label{fig-mlfa}
\end{figure}

Second, to make full use of the prior spectral and spatial expertise from the MS stream and the PAN stream, the SS stream hierarchically derives and fuses them at different levels, as shown in Fig. \ref{fig-mlfa} (c). In addition, the SS stream further extracts valuable information from the previous features to enhance the representation of the spectral-spatial features, which can be expressed as a general form  
\begin{equation}
\begin{split}
f^{k+1} = H(f^{k}) &=  f_{M}^{k}   \oplus  f_{P}^{k} \oplus f_{SS}^{k}  \\
\end{split}
\end{equation}
where $f_{M}^{k}$ , $ f_{P}^{k}$ and $f_{SS}^{k}$ denote the output of the $k$-th MS stream, the PAN stream and the SS stream, respectively, and $ \oplus $ is the element-wise addition operation.

The obvious intuition for this equation is that different types of features are combined to leverage both the  spectral and spatial information at different levels. 
This form is also unified with the motivation of Liu \cite{Liu2019}. 
This helps the SS stream effectively master a joint spectral-spatial representation in 
the hierarchical network.
By comparing the architecture with solo modeling [as illustrated in Fig. \ref{fig-mlfa} (a)], it is obvious that the hierarchical multi-level fusion can better model the fusion relationship.
Meanwhile, this also helps the architecture implement the multi-level spectral-spatial correlation. This correlation is fully exploited and strengthened to greatly suppress spectral and spatial distortions.

\subsection{Residual Spectral Attention Block}
\begin{figure}[!ht]
	\centering
	\includegraphics[width=\linewidth]{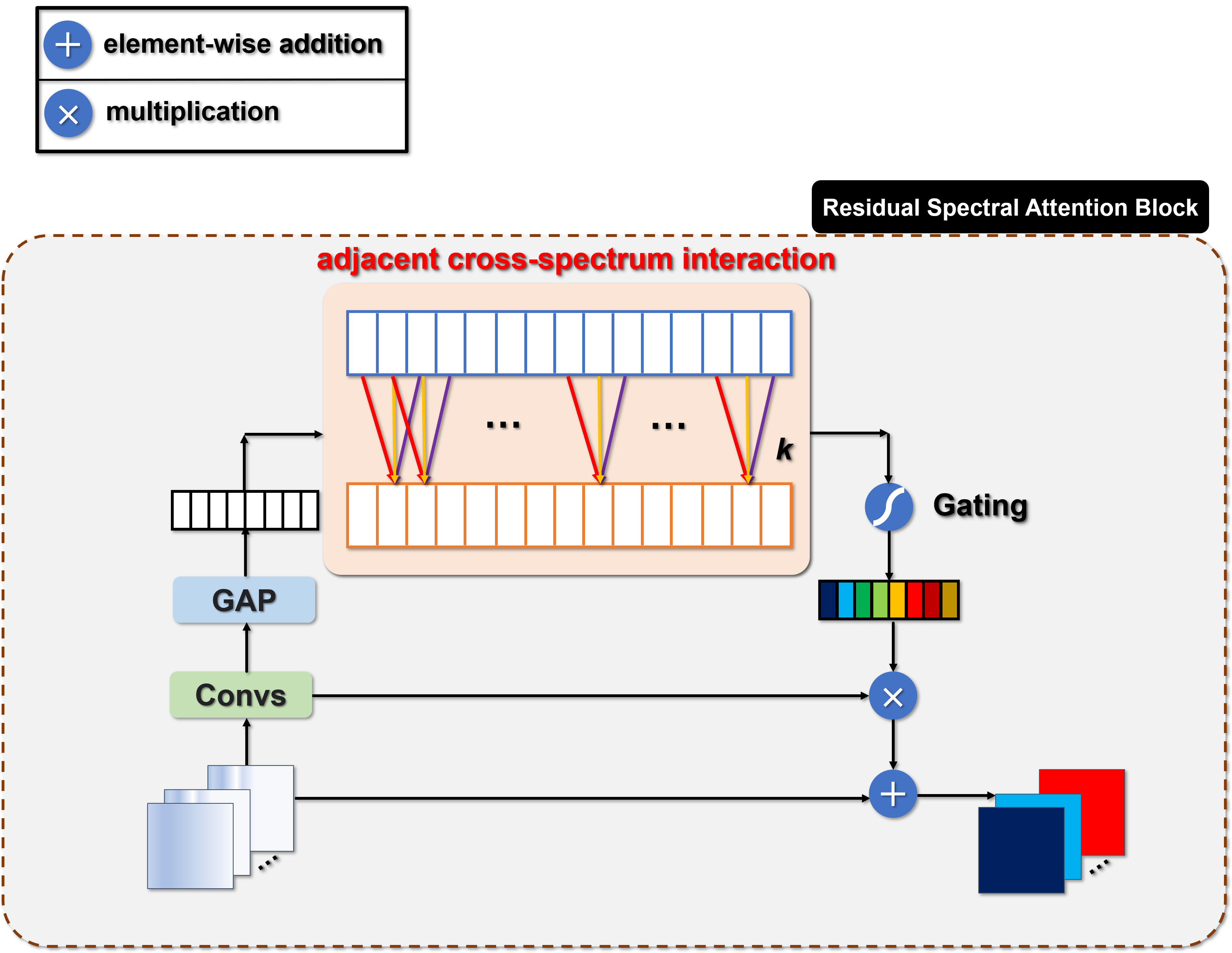}
	\caption{  The structure of RSAB, which consists of residual structure and spectral attention. } \label{fig-ESAB}
\end{figure}

There are spectral correlations between the different spectra of the MS image. We suggest that the potential correlations also exist between the adjacent spectral features. Most CNN-based pan-sharpening networks don't focus on this intrinsic characteristic of the MS image and use standard 2-D convolution to extract spectral features. This fails to effectively exploit the potential correlations, thus leading to the spectral deviations in the generated HRMS image. He \textit{et al.} \cite{He2020} employed 3-D convolution to naturally adapt the spectral dimension, but this consumes a large amount of computing resources.

To mine the correlations, we are inspired by the work of \textit{Enhanced} \textit{Channel} \textit{Attention} (ECA) \cite{Wang2020}, and design the \textit{Spectral} \textit{Attention} (SA). The structure of SA is illustrated in Fig.\ref{fig-ESAB}, which consists of the following three steps:   

First, the spectral features $f_M \in \mathbb{R}^{H \times W \times C}$ ($H, W, C$ are the height, width and number of channels of the feature maps, respectively) are compressed into the global distribution response $z \in \mathbb{R}^{ C}$ of the spectral channels through \textit{Global Average Pooling} (GAP) function. 
\begin{equation}
\begin{split}
z &= F_{GAP} ( f_M )  
\end{split}
\end{equation}
where $F_{GAP} $ denotes the Global Average Pooling function. 

Then, 1-D convolution is utilized to capture adjacent cross-spectrum interaction information by considering each spectral feature channel and its $ k $-nearest neighbors. 
\begin{equation}
\begin{split}
w &=\sigma ( {\text{Conv1D}_k} (z))  
\end{split}
\end{equation}
where $Conv1D(\cdot)_k$ and $\sigma(\cdot)$ are the 1-D convolution with kernel size is $ k $ and the sigmoid activation function, respectively. We determine the $k$ value based on the trade-off between the number of the spectral bands in the MS image  and spectral channels in the feature maps.

Finally, to represent the correlations into the network, gating mechanism is attached to form the spectral attention mask $w \in \mathbb{R}^{ C}$ for recalibrating the original features $f_M$.
\begin{equation}
\begin{split}
f_M &= f_M * w
\end{split}
\end{equation}
We embed SA into the residual structure \cite{He2016} to form the feature extraction block \textit{Residual} \textit{Spectral} \textit{Attention} \textit{Block} (RSAB) of the MS stream for extracting valuable prior spectral expertise. Spectral process and prediction are enhanced to prevent spectral deviations. The \textit{k}-th output of RSAB can be expressed as
\begin{equation}
f_{M}^{k} = SA(f_{M}^{k-1}) \oplus f_{M}^{k-1}
\end{equation}
where  $SA(\cdot)$ is SA, and $ \oplus $ is the element-wise addition operation.
\subsection{Residual Multi-scale Spatial Attention Block}
\begin{figure}[!ht]
	\centering
	\includegraphics[width=\linewidth]{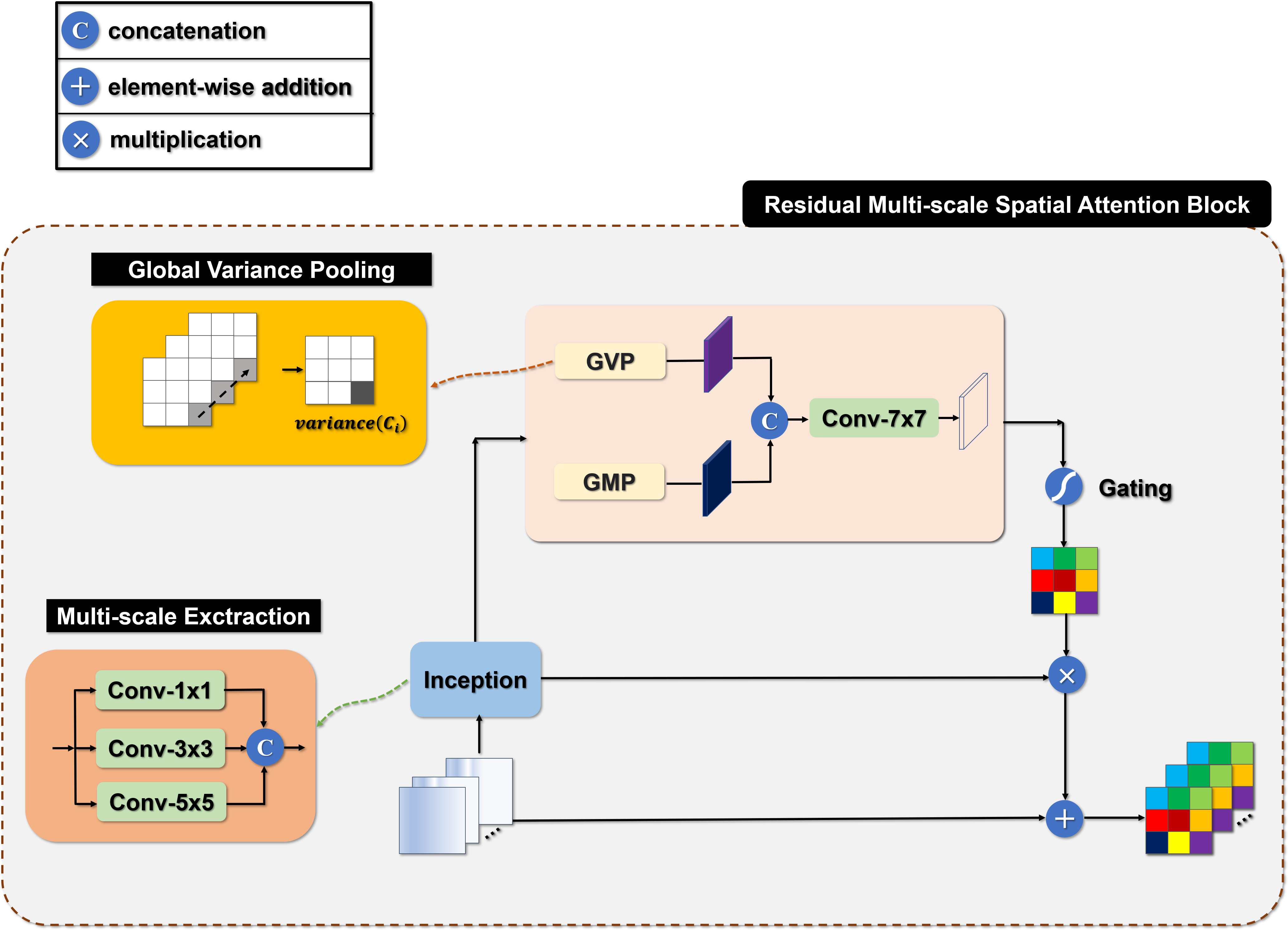}
	\caption{  The structure of RMSAB, which consists of inception structure and improved spatial attention. } \label{fig-EMSAB}
\end{figure}

There are two prominent points in the high spatial-resolution PAN image, the large scale difference and the precise spatial details.
Shao and Cai \cite{Shao2018} used a deep branch to emphasize the spatial features. Zheng \textit{et. al} \cite{Zheng2020} introduced the attention to enhance the spatial details. But they didn't give a better solution from these points.

Inspired by the Inception structure \cite{Szegedy2016}, multi-scale feature extraction (kernel sizes of 1, 3, and 5 are selected) is introduced to improve the encapsulation of spatial features, and provide the robust spatial information encoding ability. An 1 x 1 convolution is applied to adaptively fuse a series of features from different scales. Moreover, to reduce the computational cost, the symmetric n x n convolution is decomposed into asymmetric 1 x n and n x 1 convolutions. 

To further enhance the spatial details, we propose an improved spatial attention. 
Woo\textit{et. al} \cite{Woo2018} used parallel \textit{Global Max Pooling} (GMP) and \textit{Global Average Pooling} (GAP) to generate the spatial attention mask. However, for the pan-sharpening task, we aim to construct as much high-frequency information as possible. 
It is more ideal to exploit high-frequency statistics for constructing and attending the high-frequency details. It's obvious that variance statistics represents the high-frequency information in the PAN image. 
Therefore, the GMP is changed to \textit{Global Variance Pooling} (GVP). 

Given the spatial features $f_P = [f_P(1), ..., f_P(C)] \in \mathbb{R}^{H \times W \times C}$ (in this paper, $C$=64), the output of the GVP $z \in \mathbb{R}^{H \times W}$ is computed by 
\begin{equation}
z(i,j) = \frac{1}{C} \sum_{k=1}^{C} (f_P(k) - \overline{f_P})^2
\end{equation}
where $ z(i,j) $ denotes the element at the position $ (i,j) $ in $z$, and $f_P(k) $ and $\overline{f_P} $ denote the $k$-th channel of $ f_P $ and the corresponding expectation value.  

The improved spatial attention mask $ w \in \mathbb{R}^{ H \times W} $ in \cite{Woo2018} can be redefined as 
\begin{equation}
w = \sigma (Conv ( F_{GVP}(x) \ \copyright \ F_{GAP}(x) ))
\end{equation}
where 
$F_{GVP} $ and $F_{GAP} $ denote the Global Variance Pooling and the Global Max Pooling function, and $\copyright$ is concatenation operation. 

Other steps are similar to RSAB and \cite{Woo2018}. We also embed the \textit{Inception} (Inc) and the \textit{Improved Spatial Attention} (ISA) into the residual structure to form the feature extraction block \textit{Residual Multi-scale Spatial Attention Block} (RMSAB) of the PAN stream for extracting valuable prior spatial information. The spatial quality can be effectively improved. The \textit{k}-th output of RMSAB can be formulated as
\begin{equation}
f_{P}^{k} = ISA(Inc(f_{P}^{k-1})) \oplus f_{P}^{k-1}
\end{equation}
where  $Inc(\cdot)$ and $ISA(\cdot)$ are the Inception and ISA, respectively, and $ \oplus $ is the element-wise addition operation.

\subsection{Loss function}
Given the input low-resolution MS image $ I_{MS} \in \mathbb{R}^{ h \times w \times c}$ and the input high-resolution PAN image $I_{PAN} \in \mathbb{R}^{ H \times W } $, the prediction image is a high-resolution
MS image. In the pan-sharpening task, the most frequently used are $L_1$ and $L_2$ losses. Compared with $L_2$ loss, $L_1$ loss has better penalties for small errors. Another advantage of $L_1$ loss is that it is easy to get the sparse solution. Therefore, in this paper, we prefer to adopting the $L_1$ loss. Let  $y^{(i)}$ and  $f$ denote as the ground truth and the proposed MESSFN, respectively. The $L_1$ loss can be defined as
\begin{equation}
L(\Theta) = \frac {1}{N} \sum_{i=1}^{N} || y^{(i)} - f(I_{MS}^{(i)}, I_{PAN}^{(i)}; \Theta) ||_1
\end{equation}
where $\Theta$ denotes the network parameters, and $N$ is the number of image samples in the training batch.

\section{Experiments}
\label{sec-experiments}
\begin{table}[!htbp]
	\Large
	\caption{The properties of MS and PAN images for WorldView-II and GaoFen-2 datasets.}
	\centering
	\setlength{\tabcolsep}{3pt} 
	\renewcommand\arraystretch{1.3} 
	\resizebox{\linewidth}{!}{
		\begin{tabular}{lccccccc}
			\toprule
			& \multicolumn{5}{c}{ Wavelength Range/nm}       & \multicolumn{2}{c}{Spatial Resolution/m} \\ \cline{2-6}   \cline{7-8} 
			& \multicolumn{4}{c}{MS}             & \multirow{2}{*}{PAN} & \multirow{2}{*}{MS}  & \multirow{2}{*}{PAN}  \\
			\cline{2-5} 
			& NIR       & R         & G         & B         &                      &                      &                       \\
			\multicolumn{1}{c}{WorldView-\uppercase\expandafter{\romannumeral2}} & 770 – 895 & 630 - 690 & 510 - 580 & 450 – 510 & 450 – 800            & 0.5                 & 2.0                  \\
			\multicolumn{1}{c}{GaoFen-2} & 770 – 890 & 630 - 690 & 520 - 590 & 450 – 520 & 450 – 900            & 0.8                    & 3.2      \\
			\bottomrule
			
		\end{tabular}
	}  
	\label{tab-pro}	
\end{table}

\subsection{Datasets and Implementation Details}
Experiments are conducted on the WorldView-II and GaoFen-2 datasets. Both datasets contain four bands (Red, Green, Blue and Near-InfraRed), but the wavelength range and the spatial resolution of MS and PAN images are different. The specific properties are shown in Table \ref{tab-pro}.

Since the HRMS images are not available, following the Wald’s \cite{Wald1997} protocol, the original MS images are used as the ground truth. To maintain consistent spatial resolution with the real data, the original MS and PAN images need to be down-sampled by a factor $ r $, which is determined by the ratio of their spatial resolutions. We crop the source images to 64 x 64 pixels to generate about 13,000 paired simulated data samples. We divide them into 90\% as training data, and the remaining 10\% as validation data.

MESSFN is implemented in Pytorch 1.4 and runs on Ubuntu 18.04 LTS with NVIDIA TITAN X GPU.  In this paper, the $ r $ is set to 4. The network is trained for total 350 epochs with the batch size of 64. The learning rate is initialized to 0.0001 and is multiplied by 0.1 after 150 epochs to decay. The Adam \cite{Kingma2014} optimizer is used to minimize the $L_1$ loss with $ \beta_1 $=0.7 and $ \beta_2 $=0.99. We set \textit{B}=9 in the framework and will discuss it in the Section \ref{numberofB}.
\subsection{Comparison with state-of-the-art}
In this paper, eight state-of-the-art pan-sharpening methods are employed to compare with the proposed network, including PCA \cite{Shah2007a}, IHS \cite{Rahmani2010}, GS \cite{Li2016},  MTF\_GLP\_HPM \cite{Aiazzi2006a}, CNMF \cite{Karoui2016}, PNN \cite{Masi2016}, PanNet \cite{Yang2017}, RSIFNNN \cite{Shao2018}.  (In \href{https://github.com/yisun98/MESSFN}{github}, we show more fusion examples and more state-of-the-art pan-sharpening methods.)

Both real data and simulated data are used to measure the fusion performance, therefore the evaluation is performed on the two types of metrics. The reference metrics include \textit{Peak Signal to Noise Ratio} (PSNR) \cite{Nezhad2016}, \textit{Structural SIMilarity} (SSIM) \cite{Wang2004}, \textit{Spectral Angle Mapping} (SAM) \cite{Yuhas1992}, \textit{Relative Dimensionless Global Error in Synthesis} (ERGAS) \cite{Wald2002}, \textit{Correlation Coefficient} (CC) \cite{Zhu2013} and \textit{Quality-$ n $ } (in this paper, $ n $=4) \cite{Alparone2004}. These are commonly adopted to simulated data. For real data, the no-reference metrics such as spatial distortion index $ D_s $, \textit{Quality with No Reference} (QNR) \cite{Alparone2008}, and spectral distortion index $D_\lambda$ are widely recognized. 

\subsubsection{Simulated Data}
\begin{figure*}[htbp]
	\centering
	\includegraphics[width=\linewidth]{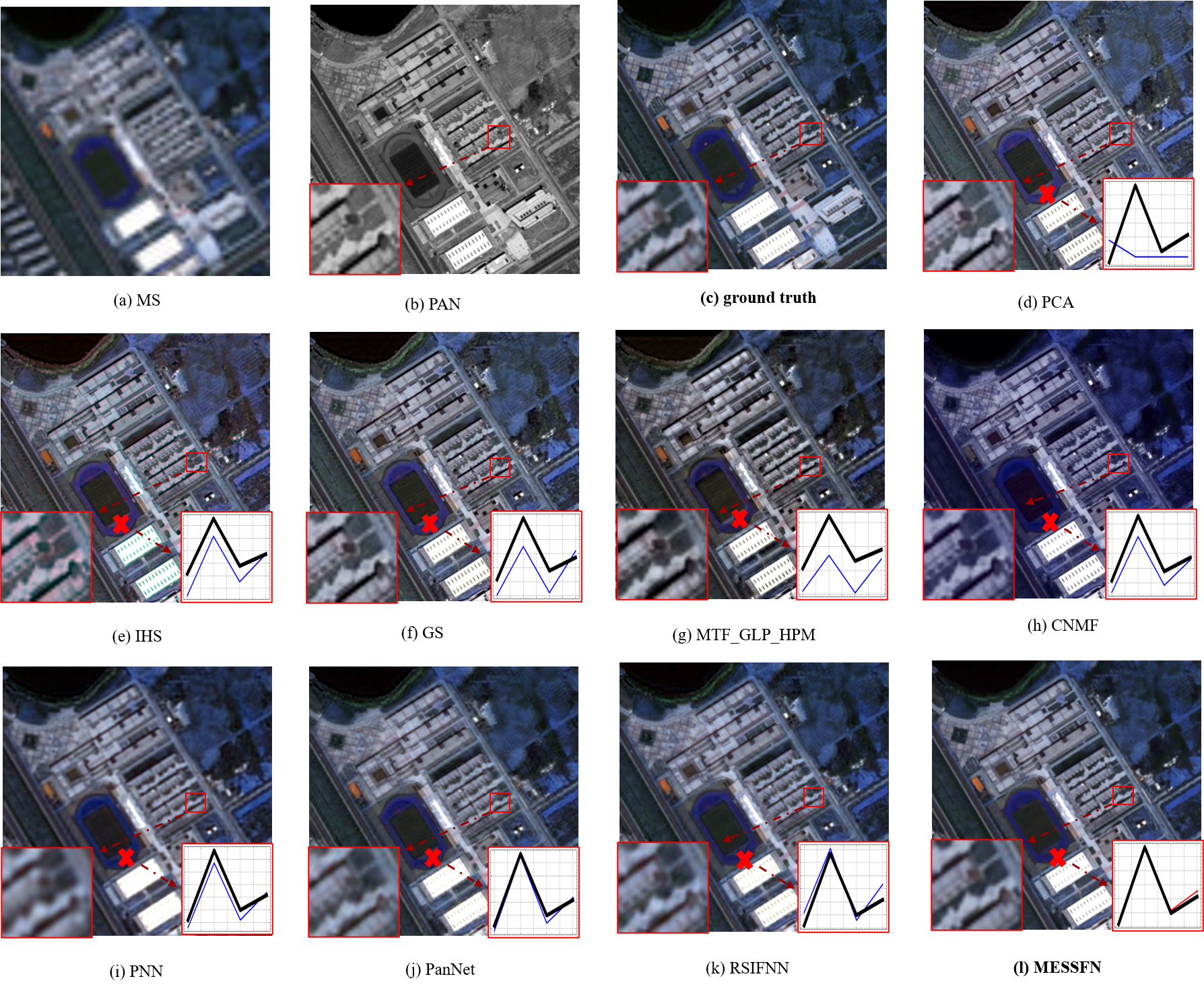}
	\caption{Visual fusion results of the simulated data on the WorldView-II dataset. The red rectangle is the emphasized region and the magnified details are on the left side. The red cross is a random coordinate point and the spectral curve based on this point is drawn on the right side. The thick black line is ground truth, and the thin blue line is the fusion result. (Please zoom in to see details.)
	}
	\label{fig-wv1}
\end{figure*} 

Fig.\ref{fig-wv1} shows the visual fusion results of the simulated data on the WorldView-II dataset. Fig.\ref{fig-wv1} (a)-(c) are the re-sampled MS image, PAN image and ground truth, respectively. Fig.\ref{fig-wv1} (d)-(l) are the fusion results of various methods. The red rectangle is the emphasized region and the magnified details are on the left side. The red cross is a random coordinate point and the spectral curve (the reflectance of the spectrum of R, G, B, NIR) based on this point is drawn on the right side.
The thick black line is ground truth, and the thin blue line is the fusion result.

Compared with the ground truth in Fig.\ref{fig-wv1} (c), visible color changes can be seen in Fig.\ref{fig-wv1} (d)-(f) and (h). At the central circular playground, the color space of Fig.\ref{fig-wv1} (j) and the spectral curve of Fig.\ref{fig-wv1} (d) entirely deviate from the reference image. This indicates that the fused images generated by PCA \cite{Shah2007a}, IHS \cite{Rahmani2010}, GS \cite{Li2016}, and CNMF \cite{Karoui2016} lose a lot of spectral information and appear serious spectral distortion. 
The buildings in Fig.\ref{fig-wv1} (g) are heavily bulged.  And Fig.\ref{fig-wv1} (i) and (j) show extremely blurred with unpleasant artifacts. 
Their obvious loss in spatial details can be easily seen in MTF\_GLP\_HPM \cite{Aiazzi2006a}, PNN \cite{Masi2016}, PanNet \cite{Yang2017}, and RSIFNN \cite{Shao2018}. 
The discrimination of features is heavily affected.
Both Fig.\ref{fig-wv1} (k) and the proposed method (l) are very close to the ground truth. Further observation of the magnified region shows that the edges of RSIFNN are not fine, and MESSFN can effectively recover these details. Meanwhile, in the lower white region, Fig.\ref{fig-wv1} (k) is brighter than ground truth and this spectral brightness is also reflected in the spectral curve. The proposed method almost exactly matches the ground truth in the R, G and B bands with excellent spectral and spatial retention. 
\begin{table}[!htbp]
	\footnotesize
	\centering
	\caption{Quantitative evaluation results of the simulated data on the WorldView-II dataset. (In \href{https://github.com/yisun98/MESSFN}{github}, we show more state-of-the-art pan-sharpening methods.)}
	\label{tab-wv1}
	\renewcommand\arraystretch{1.3} 
	\begin{tabular}{cccccccc}
		\toprule
		& PSNR / dB ($\uparrow$)             & SSIM($\uparrow$)            & SAM($\downarrow$)             & ERGAS($\downarrow$)            & CC($\uparrow$)              & Q4($\uparrow$)              &        \\ \hline
		IHS           & 33.4668          & 0.9057          & 0.0460          & 2.1528          & 0.9561          & 0.7720          \\
		PCA           & 25.3515          & 0.8856          & 0.2057          & 5.5260          & 0.9552          & 0.7568          \\
		GS            & 33.5261          & 0.9139          & 0.0434          & 2.0500          & 0.9642          & 0.7855          \\
		MTF\_GLP\_HPM & 30.1183          & 0.8725          & 0.0432          & 3.0451          & 0.9440          & 0.7351          \\
		CNMF          & 35.5258          & 0.9244          & 0.0402          & 1.6032          & 0.9774          & 0.8064          \\
		PNN           & 35.4548          & 0.9142          & 0.0391          & 1.6132          & 0.9759          & 0.7978          \\
		PanNet        & 35.9148          & 0.9208          & 0.0393          & 1.5240          & 0.9784          & 0.8017          \\
		RSIFNN        & 36.8361          & 0.9332          & 0.0383          & 1.3724          & 0.9825          & 0.8231          \\
		MESSFN(\textbf{ours})    & \textbf{41.4047} & \textbf{0.9737} & \textbf{0.0220} & \textbf{0.8133} & \textbf{0.9942} & \textbf{0.9031}
		\\ \bottomrule
	\end{tabular}
\end{table}

The quantitative evaluation results of the simulated data on the WorldView-II dataset are presented in Table \ref{tab-wv1}. The best results are bolded. The upward (downward) arrow denotes that the higher (lower) value of the metric is better. 
Table \ref{tab-wv1} reports that the proposed method is the best performance in all metrics. In particular, the advantages are noticeable in PSNR, SAM and SSIM. 
The PSNR of MESSFN is 4.6 dB higher than the sub-optimal results (RSIFNN \cite{Shao2018}). This indicates the excellent quality of the fused images of the proposed method in this paper. The improvement of SSIM and SAM are greater than the second-best method. The SSIM improves by 0.0405, which verifies the structure of the proposed MESSFN is more similar to that of the reference image.
This represents the spatial structure can be well preserved. More importantly, the SAM is reduced by 0.0163, which fully proves that the fidelity of the proposed MESSFN is optimal in the spectral domain among the various methods.

\subsubsection{Real Data}

\begin{figure*}[htbp]
	\centering
	\includegraphics[width=\linewidth]{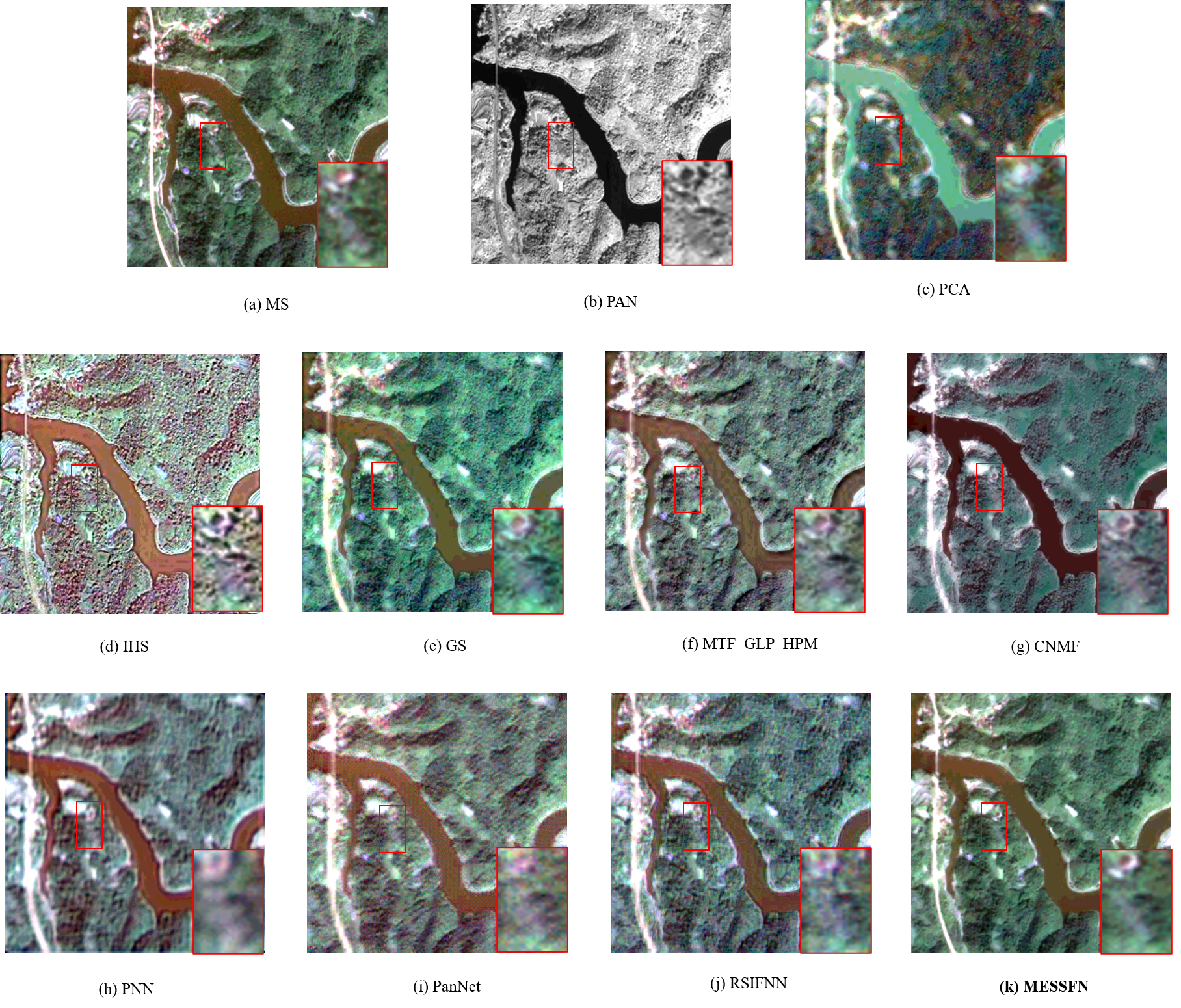}
	\caption{Visual fusion results of the real data on the GaoFen-2 dataset. The red rectangle is the emphasized region and the magnified details are on the right side. (Please zoom in to see details.)
	}
	\label{fig-gf1}
\end{figure*}

Fig.\ref{fig-gf1} shows the fusion results of the real data on the GaoFen-2 dataset. Fig.\ref{fig-gf1} (a)-(b) are the up-sampled MS image and PAN image, respectively. Fig.\ref{fig-gf1} (c)-(k) are the fusion results of different methods. Based on the results in Fig.\ref{fig-gf1}, it is evident that the spatial resolution of the MS image has improved, but there are significant differences between them. 
The spectra in Fig.\ref{fig-gf1} (c)-(e) and (g) are poor with great color contrasts.
The GS \cite{Li2016} method in Fig.\ref{fig-gf1} (e) preserves more spatial details, but the generated image is too bright. The same situation also presents in Fig.\ref{fig-gf1} (k). Fig.\ref{fig-gf1} (f) and (h) produce artifacts on contours and edges. 
This indicates that the spatial information is not well reconstructed between these methods. 
The fused images in Fig.\ref{fig-gf1} (i) and (j) are darker than the MS image. 
Interestingly, we find that in some green vegetation-covered regions (e.g., Fig.\ref{fig-gf1} and Read Data 2nd in \href{https://github.com/yisun98/MESSFN/examples/Real Data 2nd}{github}), there are different degrees of spectral loss in various methods. 
In comparison, MESSFN provides satisfactory results, which are generally consistent with the spectra of the MS image in Fig.\ref{fig-gf1} (a). We can conclude that the proposed MESSFN achieves the best fidelity in the spectrum among these methods.

\begin{table}[!htbp]
	\caption{Quantitative evaluation results of the real data on the GaoFen-2 dataset. (In \href{https://github.com/yisun98/MESSFN}{github}, we show more state-of-the-art pan-sharpening methods.)}
	\Large
	\label{tab-wv2}
	\setlength{\tabcolsep}{7mm} 
	\renewcommand\arraystretch{1.3} 
	\resizebox{\linewidth}{!}{
		\begin{tabular}{cccc}
			\toprule
			& $D_\lambda$($\downarrow$)                & $D_s$ ($\downarrow$)               & QNR  ($\uparrow$)            \\ \hline
			IHS           & 0.0801         & 0.1368         & 0.7941          \\
			PCA           & 0.0979         & 0.6742         & 0.2939          \\
			GS            & \textbf{0.0220} & 0.0548         & 0.9244          \\
			MTF\_GLP\_HPM & 0.0706         & 0.1114         & 0.8258          \\
			CNMF          & 0.0276         & 0.0548         & 0.9191          \\
			PNN           & 0.0395         & 0.0404         & 0.9217          \\
			PANNET        & 0.0673         & 0.0325         & 0.9023          \\
			RSIFNN        & 0.0605         & 0.0383         & 0.9035          \\
			MESSFN(\textbf{ours})    & 0.0296         & \textbf{0.0260} & \textbf{0.9452}
			\\ \bottomrule
		\end{tabular}
	}
\end{table}
Table \ref{tab-wv2} gives the results of the quantitative evaluation of the real data of the GaoFen-2 dataset. MESSFN has the lowest spectral distortion index among the CNN-based methods, while $D_s$ and QNR obtain the best performance in all methods. 
Compared with the CNMF method that has the highest value of $D_\lambda$, MESSFN only lags behind by 0.0046, but $D_s$ improves by a factor of two. And the improvement of QNR is 2.2\%. This demonstrates that the proposed method can retain much spatial structure information while making a perfect trade-off between the spectral and the spatial resolution, thus providing desirable fusion results. 

\subsection{Discussion and Analysis}
\subsubsection{Effect of B}
\label{numberofB}
\begin{figure}[htbp]
	\centering
	\includegraphics[width=\linewidth]{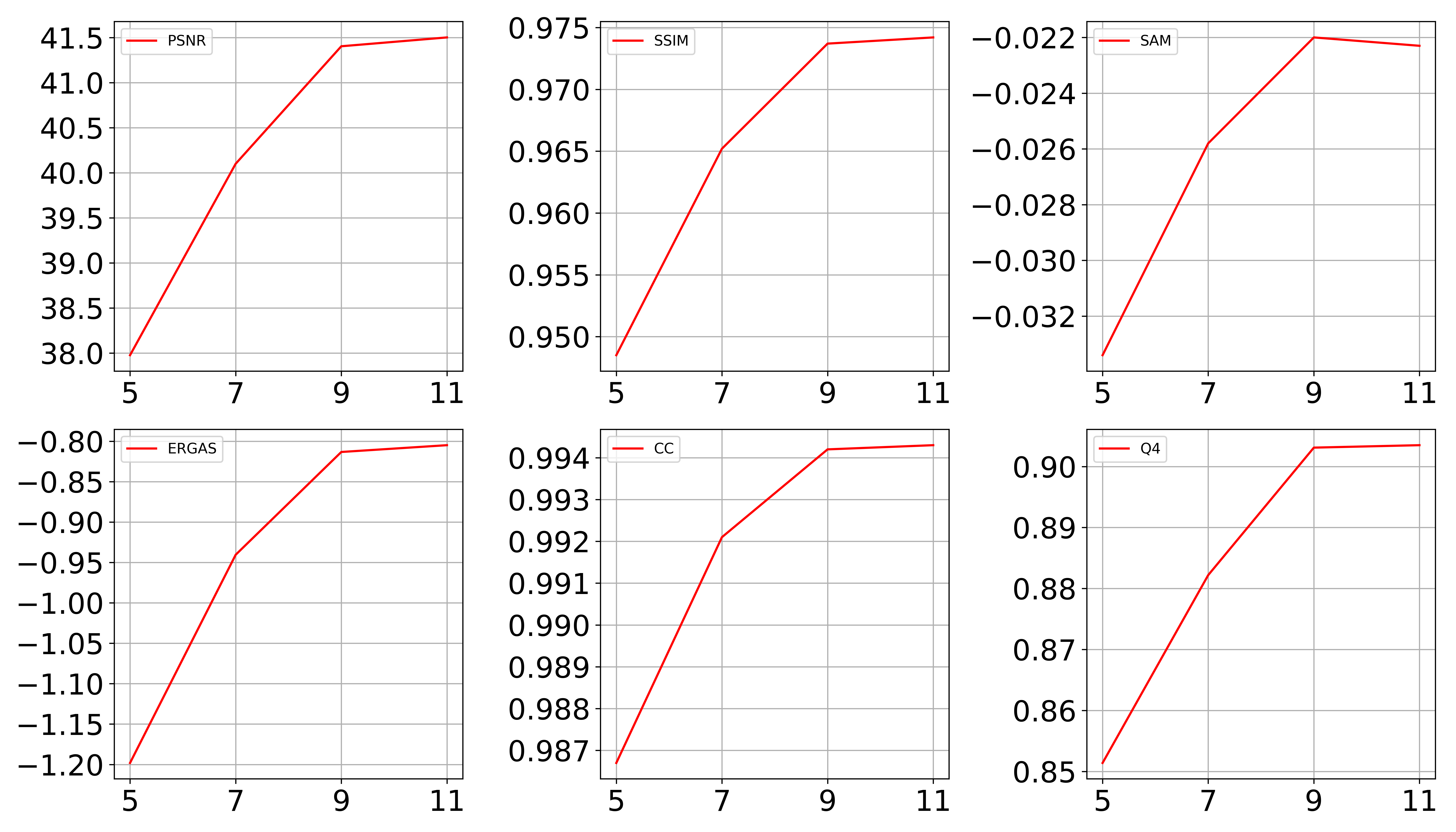}
	\caption{The fusion performance for different numbers of $ B $. For comparison, we take the negative values of the SAM and ERGAS metrics.
	}
	\label{fig-nb}
\end{figure}
Extensive works \cite{He2016, Zhang2018a} have proven that the depth is critical to the network. We cascade $ B $ feature extraction blocks in the network. In this experiment (MESSFN-B), we analyze the effect of $ B $ on the network. We set $  B $ starting from 5 to 11 with the step of 2. The quantitative results of the simulated data on the WorldView-II dataset are listed in Table \ref{tab-nb}. Fig. \ref{fig-nb} shows how the fusion performance under different numbers of $ B $. 
\begin{table}[!t]
	\caption{Quantitative evaluation results for different numbers of $B$. }
	\Large
	\label{tab-nb}
	\setlength{\tabcolsep}{3pt} 
	\renewcommand\arraystretch{1.3} 
	\resizebox{\linewidth}{!}{
		\begin{tabular}{cccccccc}
			\toprule
			& PSNR / dB ($\uparrow$)             & SSIM($\uparrow$)            & SAM($\downarrow$)             & ERGAS($\downarrow$)            & CC($\uparrow$)              & Q4($\uparrow$)              &        \\ \hline
			$B$=5        & 37.9773 & 0.9485 & 0.0334 & 1.1981 & 0.9867 & 0.8514 \\
			$B$=7           & 40.1012 & 0.9652 & 0.0258 & 0.9405 & 0.9921 & 0.8822 \\
			$B$=9          & 41.4047 & 0.9737 & \textbf{0.0220} & 0.8133 & 0.9942 & 0.9031 \\      
			$B$=11         & \textbf{41.5027} & \textbf{0.9742} & 0.0223 & \textbf{0.8049} & \textbf{0.9943} & \textbf{0.9035}
			\\ \bottomrule
		\end{tabular}
	}
\end{table}
As it can be obviously noticed that there is a significant improvement in all evaluation metrics when $ B $ changes from 5 to 9. However, PSNR, SSIM, and ERGAS are only slightly better than $ B $=9 when $ B $ reaches 11. CC and Q4 are maintained at the current numerical level, but SAM exhibits a degradation phenomenon, which is 1.36\% lower than $ B $=9. In this paper, the MESSFN with $ B $=9 obtains the most spectral and spatial structure, and the results are satisfactory in all experiments. Moreover, for more $ B $, it also brings complex network configurations and skillful training mode. Therefore, $ B $ is set to 9 in the proposed MESSFN.

\subsubsection{Effect of RSAB}
\begin{figure}[htbp]
	\centering
	\includegraphics[width=\linewidth]{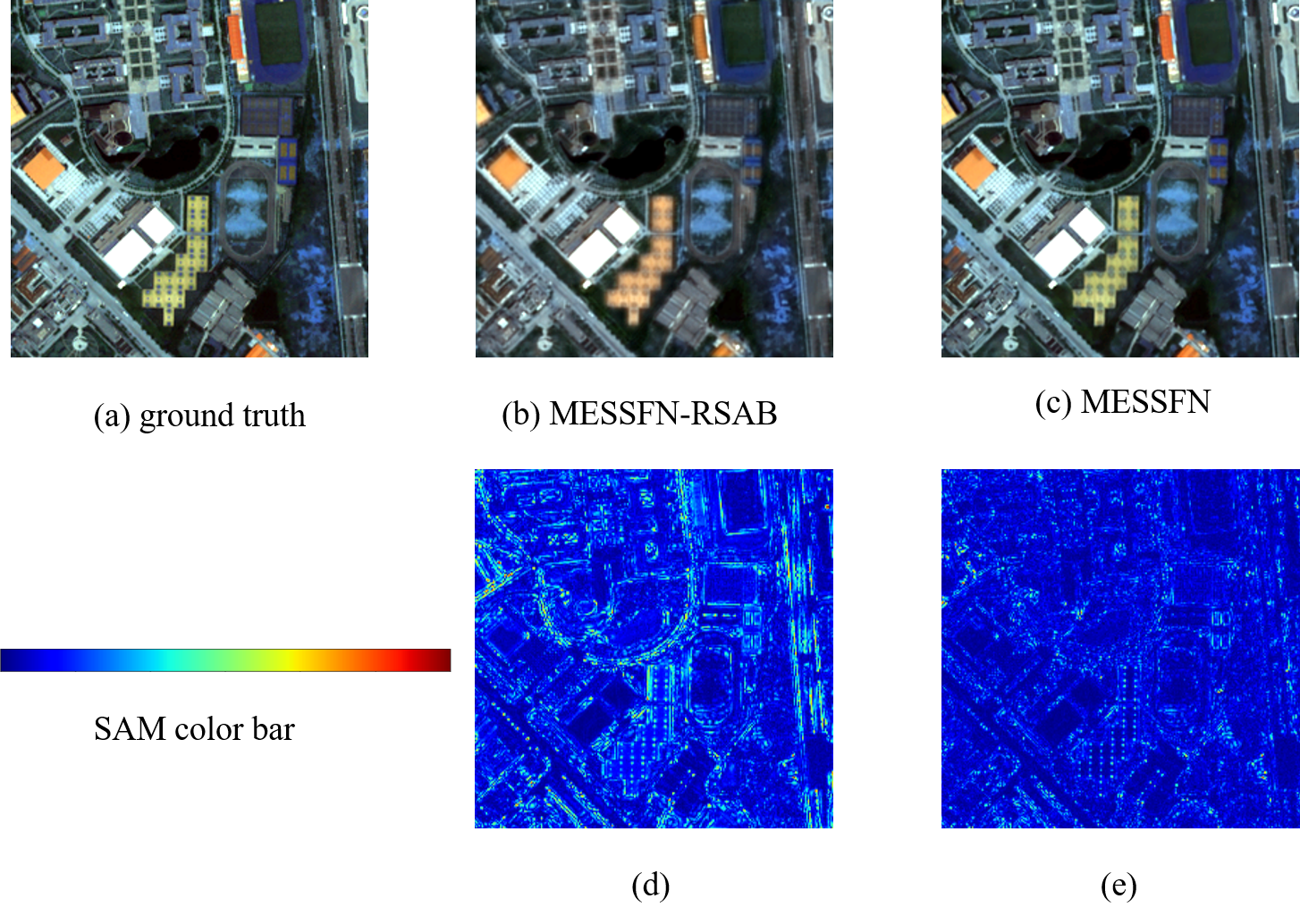}
	\caption{Experiments results of the effect of RSAB. (a) ground truth (b) MESSFN-RSAB (c) MESSFN
		(d) and (e) are the SAM images of (b) and (c), respectively. The left of the SAM color bar denotes less loss of spectral information, and the right is the opposite.
	}
	\label{fig-rsab}
\end{figure}

In this paper, RSAB is proposed to provide the spectral expertise through mining the potential spectral correlations between different spectra of the MS image. To analyze the effect of the internal structure of RSAB on the basic framework (all simple convolutions), we replace RSAB with simple convolution in the MS stream and conduct experiments (MESSFN-RSAB) on the WorldView-II dataset. The experimental results are shown in Fig. \ref{fig-rsab}.
Fig.\ref{fig-rsab} (a)-(c) are ground truth, the fusion results of MESSFN-RSAB and  MESSFN respectively. Fig.\ref{fig-rsab} (d) and (e) are the SAM images of (b) and (c). 
We make a color mapping of SAM between the fused images and ground truth to distinguish the degree of spectral distortion in different regions. The left of the color bar denotes less loss of spectral information, and the right is the opposite. According to the results in Fig.\ref{fig-rsab}, the fusion results obtained by  MESSFN-RSAB have an obvious color mutation, especially in the lower orange region. The SAM image in Fig.\ref{fig-rsab} (d) also confirms this. It’s evident from$  $ Fig.\ref{fig-rsab} (e) that the spectral angle between the MESSFN and ground truth is extremely small. This means there is no significant loss of spectrum in the fusion result of MESSFN. We can conclude that the design of the RSAB greatly suppresses the spectral deviations of the fusion results, thus achieving better spectral fidelity.

\subsubsection{Effect of RMSAB}
\begin{figure}[htbp]
	\centering
	\includegraphics[width=\linewidth]{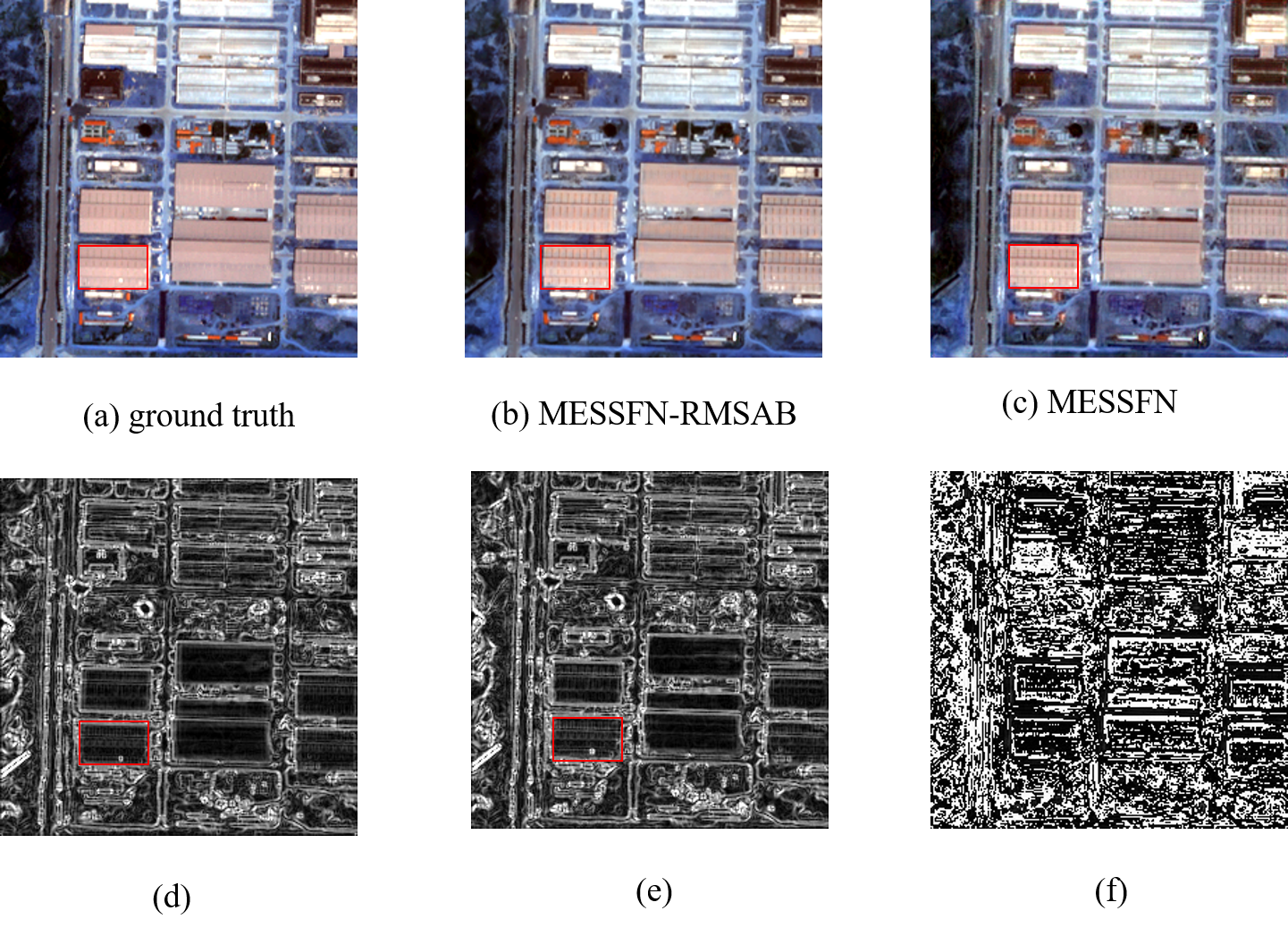}
	\caption{Experiments results of the effect of RMSAB. (a) ground truth (b) MESSFN-RMSAB (c) MESSFN. (d) and (e) are the gradient images of (b) and (c), respectively. (f) is the difference image of (d) and (e).
	}
	\label{fig-rmsab}
\end{figure}
RMSAB is designed to provide the spatial expertise based on the inherent characteristics (large-scale differences and high-spatial-resolution) of the PAN image. Similarly, we also replace RMSAB with simple convolution in the PAN stream and conduct experiments on the WorldView-II dataset. Fig.\ref{fig-rmsab} displays the experimental results.

Fig.\ref{fig-rmsab} (a)-(c) are ground truth, the fusion results of MESSFN-RMSAB and MESSFN respectively. Fig.\ref{fig-rmsab} (d) and (e) are the gradient images of (b) and (c), and Fig.\ref{fig-rmsab} (f) is the difference image of (d) and (e). It is observed that the fusion image generated by  MESSFN shows sharper edges and contours than MESSFN-RMSAB. Relying on the robust spatial coding ability of multi-scale and improved spatial attention to enhance spatial details, thus the spatial quality of the fusion results of the proposed method is significantly improved.

\subsubsection{Effect of HMFA}
\begin{figure}[htbp]
	\centering
	\includegraphics[width=\linewidth]{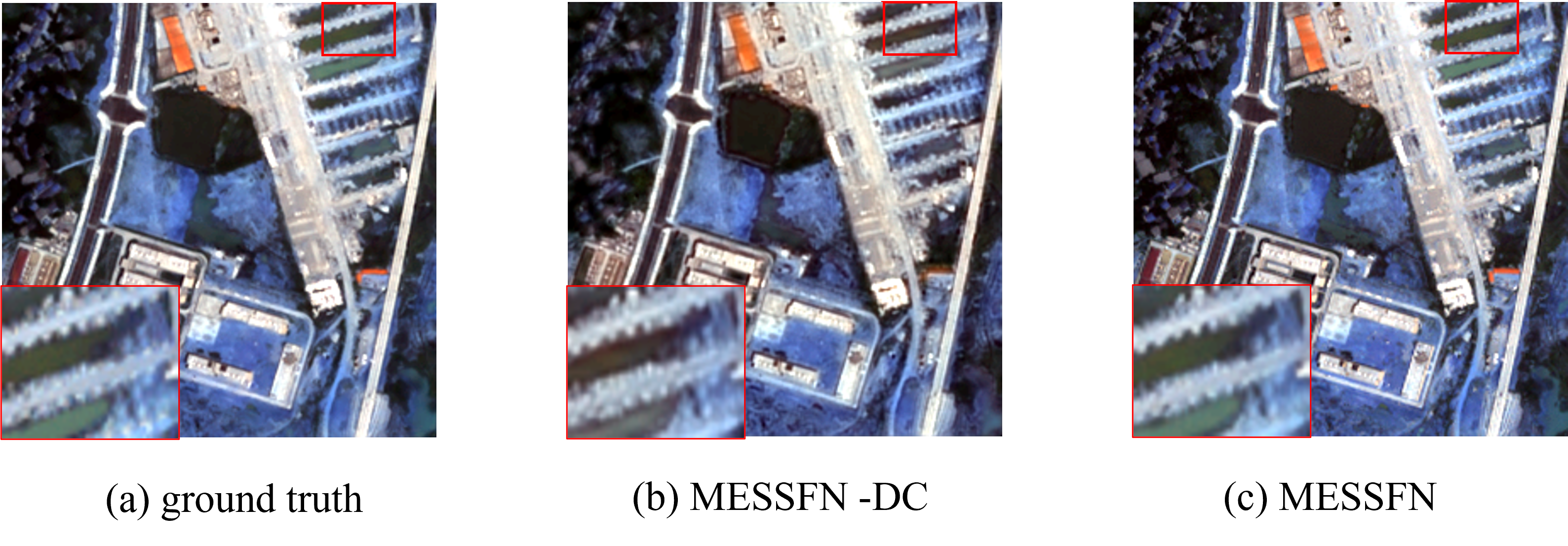}
	\caption{Visual fusion results of the effect of HMFA. (a) ground truth (b) MESSFN-DC (c) MESSFN. (Please zoom in to see details.)
	}
	\label{fig-dc}
\end{figure}

\begin{figure}[htbp]
	\centering
	\includegraphics[width=\linewidth]{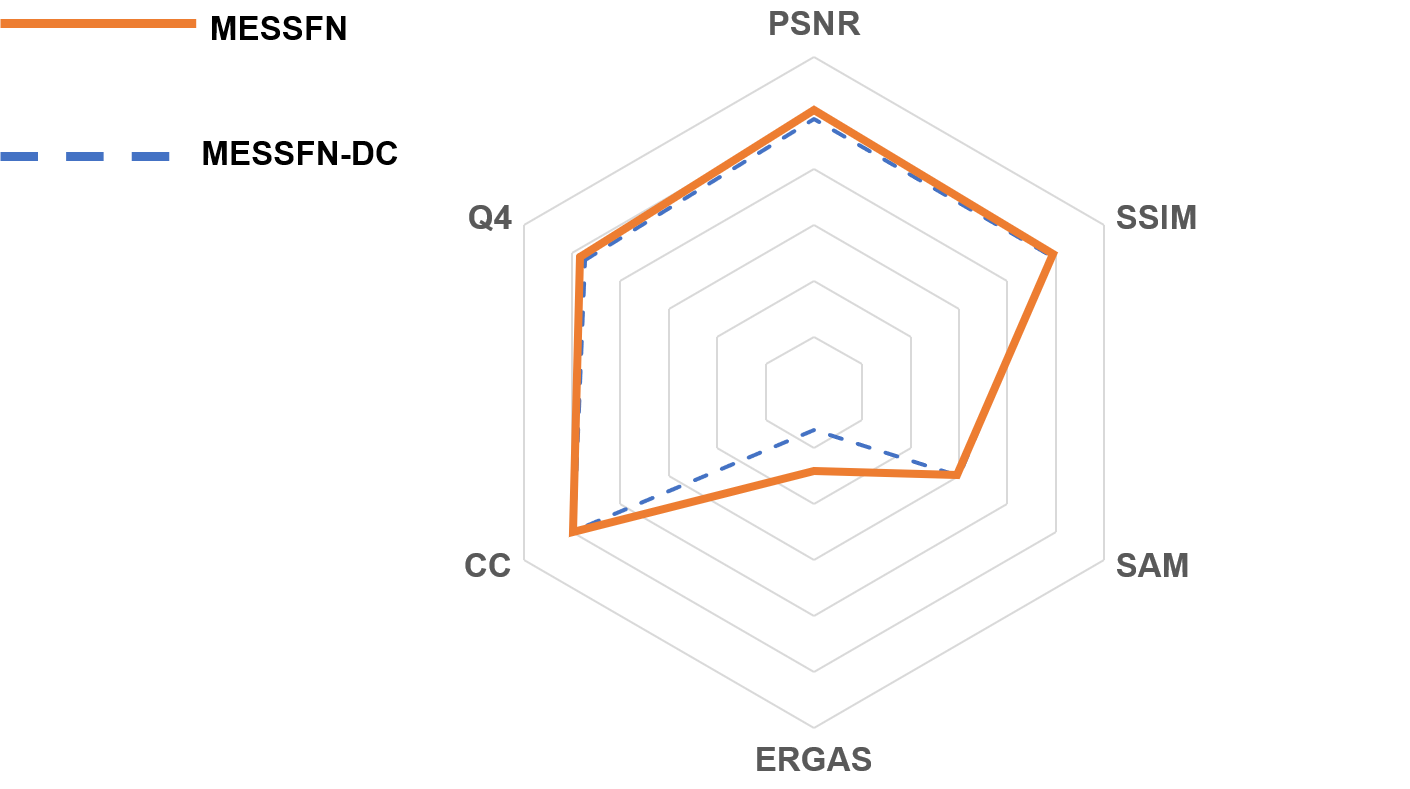}
	\caption{Experiments results of the effect of HMFA. For comparison, we take the negative values of SAM and ERGAS metrics, and rescale the value of PSNR.
	}
	\label{fig-leida}
\end{figure}
To demonstrate the effectiveness of the hierarchical multi-level fusion architecture, the most extreme approach is to completely disconnect the MS stream and the PAN stream with the SS stream. 
It can be noticed that this is essentially the same structure as RSIFNN \cite{Shao2018} and PNN \cite{Masi2016} (Only the depth of the network is different.), and the performance are confirmed in the fusion results of Fig. \ref{fig-wv1} and Fig. \ref{fig-gf1}. Therefore, in the next experiment (MESSFN-DC) on the WorldView-II dataset, 
we randomly disconnected at the $ 2 $nd, $ 5 $th,$  7 $th and $ 9 $th positions of the stream. 
The experimental results are shown in Fig. \ref{fig-dc} and Fig. \ref{fig-leida}. Fig.\ref{fig-dc} (a)-(c) are ground truth, the fusion results of MESSFN-DC and MESSFN respectively.
In the upper right region, it is clear that the MESSFN-DC is not as fine as the MESSFN in spatial. Further observation of the magnified region shows that the spectra of MESSFN-DC deviate. This similar situation does not present in the fusion results of MESSFN.
When disconnected streams in several positions, the performance of the network decreases dramatically, which is evident in Fig. \ref{fig-leida}. SSIM and SAM decrease by 2.53\% and 34.26\%, respectively. 
Few connections can't ensure to sufficiently strengthen the correlation between the spectral information and spatial information at multiple levels, and it will also cause the SS stream to lose the complete representation of spectral-spatial in the hierarchical network.

\section{Conclusion}
\label{sec-conclusion}
We propose a Multi-level and Enhanced Spectral-Spatial end-to-end Fusion Network MESSFN for pan-sharpening. First, we carefully design a Hierarchical Multi-level fusion Architecture (HMFA) in the network. In this architecture, a novel Spectral-Spatial (SS) stream is established. The SS stream hierarchically derives and fuses the multi-level prior spectral and spatial expertise from the MS stream and the PAN stream during the feed-forward procedure of the network. This practice has two main advantages: on the one hand, this helps the SS stream master a joint spectral-spatial representation in the hierarchical network for better modeling the fusion relationship. On the other hand, the multi-level spectral-spatial correlation is fully exploited and strengthened. 
Further, we should note that it is important to provide the superior expertise. Therefore we specially develop two feature extraction blocks based on the intrinsic characteristics of the MS image and the PAN image, respectively. In the MS stream, Residual Spectral Attention Block (RSAB) is proposed to mine the potential spectral correlations between different spectra of the MS image through adjacent cross-spectrum interaction along the spectral dimension. Spectral process and prediction are enhanced to prevent spectral deviations. While in the PAN stream, Residual Multi-scale Spatial Attention Block (RMSAB) is leveraged to capture multi-scale information using inception structure, and an improved spatial attention is advocated to reconstruct precise high-frequency details from the PAN image through variance statistics. The spatial quality can be effectively improved. 
Experiments on WorldView-II and GaoFen-2 datasets demonstrate that the proposed method achieves the best results in quality evaluation
among various state-of-the-art methods, and has satisfactory fusion results in visual perception. MESSFN can better achieve dual fidelity in spectral and spatial domains. In the next work, we will continue to explore more efficient fusion networks based on this architecture.

\section*{Acknowledgements}
This work was supported in part by the National Natural Science Found for Distinguished Young Scholars under Grant 61825603, 
and in part by the State Key Program of National Natural Science of China under Grant 61632018.








\bibliography{ref}

\end{document}